\documentclass{article}
\usepackage{ijcai20}

\usepackage{times}
\usepackage{soul}
\usepackage{url}
\usepackage[hidelinks]{hyperref}
\usepackage[utf8]{inputenc}
\usepackage[small]{caption}
\usepackage{graphicx}
\usepackage{amsmath}
\usepackage{amsthm}
\usepackage{booktabs}
\usepackage{algorithm}
\usepackage{algorithmic}
\graphicspath{{diagrams/}}
\usepackage{textcomp}

\usepackage{mathtools}
\DeclarePairedDelimiter\abs{\lvert}{\rvert}

\DeclarePairedDelimiter\round{\lfloor}{\rceil}

\title{Learning Accurate Integer Transformer Machine-Translation Models}

\author{
    Ephrem Wu
    \affiliations
    Xilinx, Inc., San Jose, USA
    \emails
    ephrem.wu@xilinx.com
}

\begin{document}

\maketitle
%%%%%%%% Quantization of Transformer %%%%%%%%%%%%%%%%%
\begin{abstract}
We describe a method for training accurate Transformer machine-translation
models to run inference using 8-bit integer (INT8)
hardware matrix multipliers, as opposed to the more costly
single-precision floating-point (FP32) hardware.
Unlike previous work, which converted only 85 Transformer
matrix multiplications to INT8, 
leaving 48 out of 133 of them in FP32 because of unacceptable
accuracy loss, 
we convert them \textit{all} to INT8 without
compromising accuracy.
Tested on the \textit{newstest2014} English-to-German translation task,
our INT8 Transformer Base and Transformer Big models
yield BLEU scores that are 99.3\% to 100\% relative to those of
the corresponding FP32 models.
Our approach converts all matrix-multiplication tensors
from an existing FP32 model into INT8 tensors 
by automatically making range-precision trade-offs
during training.
To demonstrate the robustness of this approach,
we also include results from INT6 Transformer models.
\end{abstract}

\section{Introduction}
We report a method for training accurate yet compact Transformer
machine-translation models \cite{DBLP:conf/nips/VaswaniSPUJGKP17}. 
Specifically, we aim these models at hardware with
8-bit integer (INT8) matrix multipliers. 
Compared to single-precision floating-point (FP32) matrix multiplications,
INT8 matrix multiplications not only reduce both storage and bandwidth
four times, but
they also consume 15 times less energy \cite{horowitz:computing_energy_problem}.
We therefore have two goals: 
1) to convert \textit{all} matrix multiplications from FP32 to INT8, and 2) to
maintain translation accuracy relative to the FP32 model.

The Transformer model has proven to be a powerful attention-based model for
machine translation \cite{Bahdanau2015},
and has inspired much follow-up research in this area.
For example, 
the term ``Transformer'' appears 56 times in the findings of
the 2018 Conference of Machine Translation (WMT18)
\cite{bojar-etal-2018-findings}
and 105 times in WMT19
\cite{barrault-etal-2019-findings}.
The Transformer model in \cite{DBLP:conf/nips/VaswaniSPUJGKP17}
comes in two forms, Transformer Base, which has 61 million parameters,
and Transformer Big, which has 210 million parameters.
These model sizes are large compared to the convolutional neural network
benchmark ResNet-50 \cite{DBLP:conf/cvpr/HeZRS16}, a 25M-parameter model,
but still small compared to OpenAI GPT-2 models \cite{radford2019language},
a family of Transformer models with 117M, 345M, 774M, and 1.6B parameters.
Transformer model sizes show no signs of reduction.
For instance,
\cite{NIPS2019_8305} reports multi-lingual machine-translation results even for larger
Transformer models for up to 6 billion parameters.
It is therefore useful to explore techniques for reducing parameter representation
costs, for instance, going from FP32 to INT8, rather than holding out for
smaller Transformer architectures.

\section{Related Work}
We draw inspiration from three papers on quantizing machine-translation models
to INT8.
The first paper, \cite{DBLP:journals/corr/WuSCLNMKCGMKSJL16}, was published
before the Transformer \cite{DBLP:conf/nips/VaswaniSPUJGKP17}.
The authors of this paper quantized parameter (weight) tensors
but not non-parameter tensors in a range-preserving fashion.
The second paper, \cite{DBLP:conf/aclnmt/Junczys-Dowmunt18}, reported
INT8 Transformer Big English-to-German translation results but
did not report any results for INT8 Transformer Base.
The third paper, \cite{DBLP:journals/corr/abs-1906-00532}, did the opposite:
it reported INT8 Transformer Base English-to-German translation results but not for
Transformer Big.

Before the Transformer paper was published,
\cite{DBLP:journals/corr/WuSCLNMKCGMKSJL16}
described a quantization-aware training method for an
attention-based LSTM neural machine translator.
They treated parameter tensors differently from non-parameter tensors.
In particular, they quantized parameter tensors to INT8 in a range-preserving fashion.
For non-parameters, they treated logits differently from other non-parameter tensors.
Specifically, they clipped logits to $[-25.0, 25.0]$, and
they clipped the rest of the non-parameter tensors to $[-8.0, 8.0]$ and
annealed them to $[-1.0, 1.0]$ by the end of the training.
In the LSTM module,
matrix multiplication operands were 8-bit integers, and accumulators were 16-bit integers.
All other operations in the LSTM module were 16-bit operations.
The softmax function and the attention mechanism remained as floating-point operations.
These authors reported an English-to-German translation BLEU score of 24.61 (\textit{newstest2014}).
Similarly, we quantized parameters to 8-bit integers in a range-preserving manner.
We attempted to clip floating-point non-parameter tensors before uniform quantization,
but observed that clipping after rounding was simpler and yielded
accurate Transformer models.
Furthermore, we did not manually select clipping ranges for non-parameters.
Our training method automatically adjusts clipping ranges to make
range-precision trade-offs.

On the basis of a literature overview, we believe that Microsoft's Marian team
was the first to publish \textit{newstest2014} English-to-German translation
BLEU scores using integer Transformer models
\cite{DBLP:conf/aclnmt/Junczys-Dowmunt18}.
With a beam size of 1,
the FP32 Transformer Big model achieved a BLEU score of 28.1, and
the 8-bit model scored 27.5.
The BLEU score of the 8-bit Transformer Base model was absent, but
the 16-bit model yielded 27.4, which is the same as that of the FP32 model.
These authors did not have to retrain the 16-bit Transformer Base model;
FP32 parameters and non-parameter tensors were scaled down by 1024 before
16-bit uniform quantization to prevent overflow.
The 8-bit Transformer Big model, however, required retraining.
Matrix multiplication input tensors (both parameters and non-parameters)
were clipped to the range $[-2, 2]$ to maximize BLEU scores
before 8-bit uniform quantization.
Similarly, we retrained Transformer Big to obtain an INT8 model.
Although the INT8 model in \cite{DBLP:conf/aclnmt/Junczys-Dowmunt18}
exhibits a 2.13\% drop in BLEU scores relative to the original FP32 model,
our INT8 model shows only 0.334\% and 0.685\% drops relative to our FP32 model
for case-insensitive (uncased) and case-sensitive (cased) results, respectively.

\cite{DBLP:journals/corr/abs-1906-00532} used calibration
to obtain an INT8 Transformer Base model from a pre-trained FP32 model.
Before summarizing this approach, a distinction should be made between 
two types of layers that multiply matrices because
these authors only quantized a \textit{subset} of one of these layer types.
These two layer types are called \texttt{dense} vs. \texttt{matmul}.
Although a \texttt{dense} layer multiplies a trainable parameter matrix by
a non-parameter matrix (e.g., a matrix of activations),
a \texttt{matmul} layer multiplies two non-parameter matrices
(e.g., a query matrix and a key matrix).
In both Transformer Base and Transformer Big,
there are 97 \texttt{dense} layers and 36 \texttt{matmul} layers.
The \texttt{matmul} layers compute attention weights, a key feature of the attention-based
Transformer model, and we also convert these layers to INT8.
To avoid unacceptable accuracy loss, these authors left 48 out of 133 matrix multiplications as FP32 operations, namely, 12 out of 97 \texttt{dense} layers and all 36 \texttt{matmul} layers.
They calibrated the FP32 model for conversion to INT8 using 600 out of 3003 sentences of various lengths from the validation dataset.
For each \texttt{dense} layer, they used KL divergence to limit the floating-point range
before uniform quantization.
Using symmetric 8-bit uniform quantization, this method achieved a BLEU score of 27.30
for the \textit{newstest2014} English-to-German translation task,
a drop of 0.38 points or 1.4\% relative to the FP32 model.
There were no accuracy results for Transformer Big.

We obtained accurate 8-bit Transformer Base and Transformer Big models
for English-to-German translation on \textit{newstest2014}.
In addition, we separately reported the BLEU scores for both cased and uncased
scenarios.
We converted all 133 matrix-multiplications, 97 \texttt{dense} layers and 36 additional
\texttt{matmul} layers in the attention module, into 8-bit integer operations.
We did so without having to manually decide which layers should be converted to INT8 and which
should not.
We let our models learn optimal range-precision trade-offs for all non-parameter tensors.
The resulting models yielded BLEU scores that were 99.3\% to 100\% relative to
to those from the FP32 reference models.

\section{Integer Matrix Multiplication}
\begin{figure}[ht]
%\vskip 0.2in
\begin{center}
\centerline{\includegraphics[width=\columnwidth]{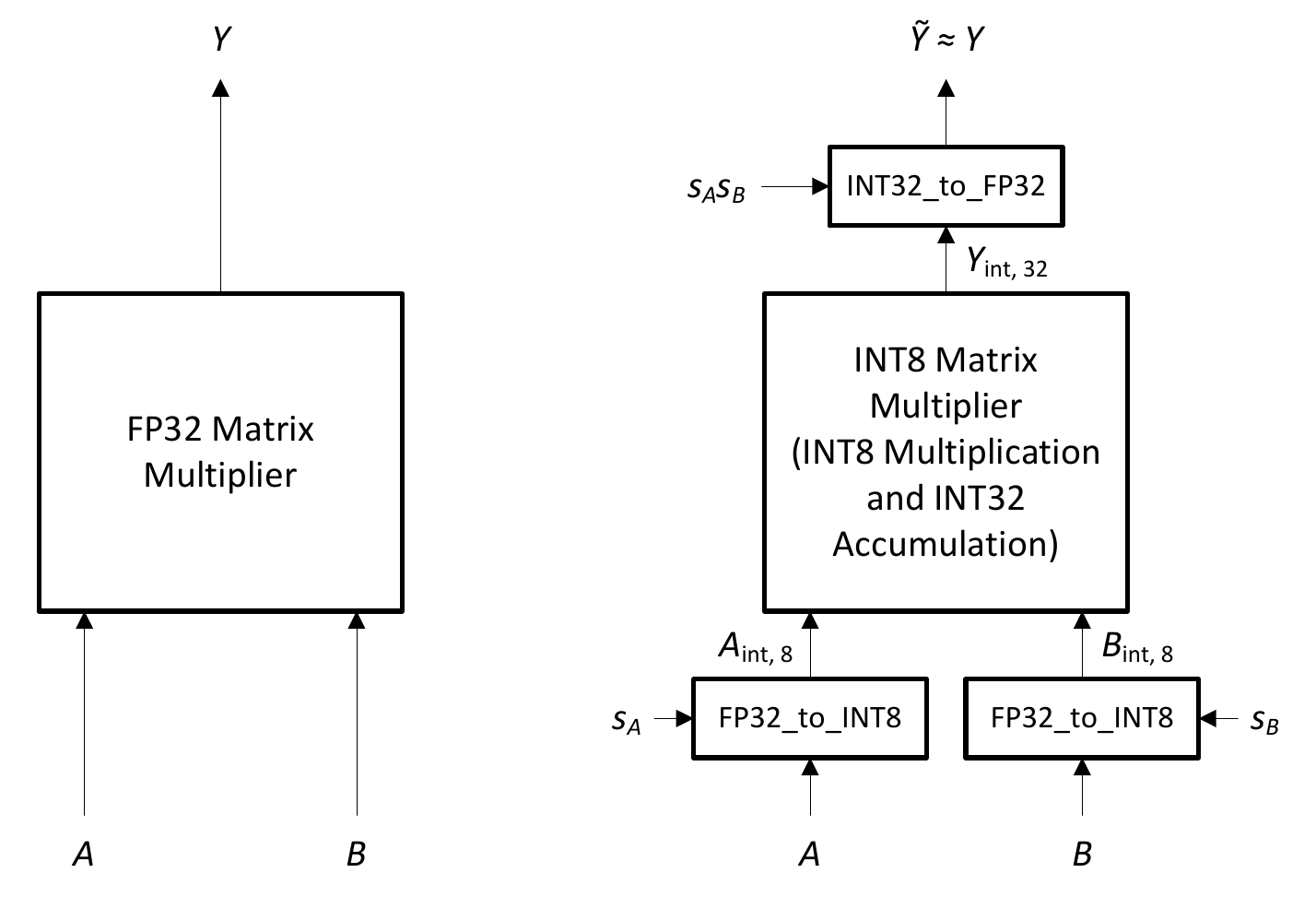}}
\caption{Applying floating-point operands to an integer matrix multiplier.
The FP32 matrix multiplier on the left computes $Y \leftarrow AB$.
The integer matrix multiplier on the right approximates
$Y$ as $s_A s_B Y_{\text{int},32}$, 
where $s_A$ and $s_B$ are the threshold scalars for $A$ and $B$, respectively,
such that
$A \approx s_A  A_{\text{int}, 8}$ and
$B \approx s_B  B_{\text{int}, 8}$.}
\label{fig:matmul}
\end{center}
\vskip -0.2in
\end{figure}

To use an integer matrix multiplier to approximate floating-point matrix
multiplication, we approximate a floating-point
tensor $\mathcal{X}$ as the product of a floating-point 
\textit{threshold scalar}\footnote{The threshold
scalar may be limited to an integer power of 2 so that multiplication 
by an integer matrix becomes an arithmetic shift operation.}
$s_\mathcal{X}$
and an integer tensor $\mathcal{X_{\text{int}}}$:
  \begin{equation}
\mathcal{X} \approx s_\mathcal{X} \mathcal{X}_\text{int}. \label{eq_i2f}
\end{equation}
Sections \ref{h:range_preserving} and \ref{h:range_precison} will discuss how a model learns its threshold scalars.
In hardware, we surround
an integer matrix multiplier with data-type converters whose
cost is amortized over the matrix multiplier array (Fig.\ \ref{fig:matmul}).

To explicityly indicate data types in tensor variables, we use a subscript
``int,'' ``uint,'' or ``fp'' for signed integers, unsigned integers,
or floating-point numbers, respectively. 
This subscript is optionally followed by data-type width in bits.
Without a data-type subscript, tensor elements by default are considered
to be IEEE single-precision floating-point (FP32) numbers.
See Fig.\ \ref{fig:matmul} for an example.

Converting an integer tensor into a floating-point tensor given a threshold scalar
is straightforward with Eq.\ \ref{eq_i2f},
but starting with a floating-point tensor $\mathcal{X}$,
which is the case after we have trained a floating-point model,
both the threshold scalar $s_{\mathcal{X}}$ and the integer tensor
$\mathcal{X}_\text{int}$ are unknown.
The question is how the threshold scalar should be set.
If the distribution of $\mathcal{X}$ is such that
saturating more of the tails will reduce training loss, 
then we should pick a sufficiently small threshold scalar
to clip more of the tails.
The threshold scalar is therefore a knob for choosing between 
range and precision.
Experimentally, we determined that in order to stabilize training,
we should preserve the range for parameter (weight and bias) tensors
and learn the threshold scalars for non-parameter tensors.

Specifically, given $b$ bits for encoding a signed integer,
we use the integer range
$[-p, p]$, where $ p = 2^{b-1}-1$:
\begin{equation}
\mathcal{X}_{\text{int}, b}  =
  \text{clip}(\left\lfloor{\mathcal{X}/s_\mathcal{X}}\right\rceil; -p, p). \label{eq:signed}
\end{equation}
If all elements in $\mathcal{X}$ are non-negative,
we use the range $[0, 2^b-1]$:
\begin{equation}
\mathcal{X}_{\text{uint}, b}  =
  \text{clip}(\left\lfloor{\mathcal{X}/s'_\mathcal{X}}\right\rceil; 0, 2^b-1),
  x \ge 0, x\in \mathcal{X}. \label{eq:unsigned}
\end{equation}
In Eqs.\ \ref{eq:signed} and \ref{eq:unsigned}
we denote bankers' rounding of a floating-point tensor $\mathcal{X}$ as
$\left\lfloor\mathcal{X}\right\rceil$ and
define $\text{clip}\left({\mathcal{X}; n, p}\right) \equiv 
\min{(\max{(\mathcal{X}, n)}, p)}$.
%\max{(\min{(\mathcal{X}, p)}, n)}$.
Note that, unlike it is done in \cite{DBLP:journals/corr/WuSCLNMKCGMKSJL16} and
\cite{DBLP:conf/aclnmt/Junczys-Dowmunt18},
these equations do not clip the original FP32 tensor $\mathcal{X}$.
Clipping occurs only after the rounding of $\mathcal{X}/s_{\mathcal{X}}$ or
$\mathcal{X}/s'_{\mathcal{X}}$ to integers.
Experimentally, we observed that clipping FP32 tensors directly is detrimental
to accuarcy,
so the goal is to find per-tensor threshold scalars that maintain accuracy.
To this end, we  separate threshold scalars into range-preserving
vs. non-range-preserving ones.
We call a threshold scalar that does not clip the distribution
\textit{range-preserving}.
Such threshold scalars are dervied from the tensor range,
and those that are non-range-preserving are learned.

\subsection{Range-Preserving Quantization}\label{h:range_preserving}
Threshold scalars $s_\mathcal{X}$ and $s'_\mathcal{X}$ in
Eqs.\ \ref{eq:signed} and \ref{eq:unsigned} represent
range-precision trade-offs.
They are range-preserving when used for signed elements,
\begin{equation}
s_\mathcal{X} =  \max_{x \in \mathcal{X}}{|x|}/p, \label{eq:s_X}
\end{equation}
and for unsigned elements,
\begin{equation}
s'_\mathcal{X}  = \max_{x \in \mathcal{X}}{|x|}/(2^b-1), x \ge 0, x\in \mathcal{X}. \label{eq:sp_X}
\end{equation}
In Transformer, we compute range-preserving threshold scalars for the weight and bias tensors in all 97
\texttt{dense} layers.

\subsection{Learned Range-Precision Trade-offs}\label{h:range_precison}
We train threshold scalars for non-parameter tensors to achieve optimal
range-precision trade-offs \cite{DBLP:journals/corr/abs-1903-08066}.
If precision is more important than range in minimizing loss,
training reduces the threshold scalar.
Conversely, if range is more important than precision,
training increases the threshold scalar.

In Transformer, we train the threshold scalars for
the remaining operands in all 97 \texttt{dense} layers,
as well as the queries and key-value pairs in the attention modules.
In a \texttt{dense} layer that computes $XW$, the weight operand $W$ uses the
range-based scalar $s_W$ as computed by Eq.\ \ref{eq:s_X},
and the other operand $X$ has its threshold scalar $s_X$ learned.
During inference, the integer weight matrix $W_\text{int}$ and the product of
threshold scalars $s_X s_W$ become constant tensors:
\begin{equation*}
P = X W \approx s_X s_W X_\text{int} W_\text{int}.
\end{equation*}
Specifically, the integer matrix multiplier receives two floating-point
constants during inference: $1/s_X$ to convert $X$ to integers and
$s_Xs_W$ to convert the integer matrix product back into the floating-point format.
We compute the integer weight matrix $W_\text{int}$ offline using
Eqs.\ \ref{eq:signed} and \ref{eq:s_X}.
During inference, the accelerator computes
\begin{align*}
X_{\text{int}} &= \text{clip}(\round{X/s_X}; -p, p), \\
P_{\text{int}} &= X_{\text{int}} W_{\text{int}}, \text{and} \\
P              &\approx s_Xs_W P_\text{int}.
\end{align*}

\subsubsection{Custom Gradients}
Following \cite{DBLP:journals/corr/abs-1903-08066},
we compute custom gradients to train threshold scalars
for range-precision trade-offs.
On the basis of Eq.\ \ref{eq:signed},
the element-wise operation that produces a floating-pointer tensor $\mathcal{Y}$
with quantization noise from the input floating-point tensor $\mathcal{X}$ is
\begin{equation*}
\mathcal{Y} = s_\mathcal{X} \text{clip}(\round{\mathcal{X}/s_\mathcal{X}}; -p, p),
s_\mathcal{X} \ge \epsilon > 0,
\end{equation*}
where the threshold scalar $s_\mathcal{X}$ is constrained to be positive.
Let $y \in \mathcal{Y}$ be dependent on some $x \in \mathcal{X}$.
The local gradient with respect to the floating-point tensor is
a straight-through estimator (STE) \cite{DBLP:journals/corr/BengioLC13}.
\begin{equation}
\frac{\partial y}{\partial x}=
\begin{cases}
    1, & \abs{\round{x/s_\mathcal{X}}} \le p \\
    0, & \abs{\round{x/s_\mathcal{X}}} > p.
\end{cases}
\label{dy_dx}
\end{equation}
For stability reasons,
the threshold scalar is
trained according to the log of this scalar, i.e., $z = \log_2{s_\mathcal{X}}$
(see Appendix B in \cite{DBLP:journals/corr/abs-1903-08066}).
Again, using STE, the local gradient with respect to $z$ is
\begin{align}
\begin{split}
\frac{\partial y}{\partial z} &=
  \frac{\partial y}{\partial s_\mathcal{X}} \cdot 
  \frac{\partial s_\mathcal{X}}{\partial z} \\
  &= s_\mathcal{X}\ln{(2)}
\begin{cases}
    \round{x/s_\mathcal{X}} - x/s_\mathcal{X},
  & \abs{\round{x/s_\mathcal{X}}} \le p \\
    p\cdot\text{sgn}(x),
  & \abs{\round{x/s_\mathcal{X}}} > p.
\end{cases}\label{dy_db}
\end{split}
\end{align}
In TensorFlow, we implemented these custom gradient functions
using the \verb+tf.custom_gradient+ decorator in Python.
Specifically, range-preserving tensors use only Eq.\ \ref{dy_dx}
and not Eq.\ \ref{dy_db}
since their threshold scalars in each step are always calculated
according to either Eq.\ \ref{eq:signed} or Eq.\ \ref{eq:unsigned}.

\subsubsection{Attention Mechanism}
The attention mechanism in the Transformer is worthy of note.
Each soft look-up in an attention module has the form
$A = \text{softmax}(QK^T/\sqrt{d_k})V$,
where $d_k$ is the width of $K$.
None of the matrices $Q$, $K$, or $V$ are constant during inference,
so we use only non-range-preserving scalars for all matrix multiplication
operands to compute $A$ from the three matrices.
Given the floating-point tensors $Q$, $K$, and $V$ during training,
we encode each integer element with $b$ bits, and
train the threshold scalars $s_Q$, $s_K$, $s_U$, and $s_V$ for inference.
Specifically,
\begin{align*}
Y &= QK^T/\sqrt{d_k} 
  \approx \frac{s_{Q}s_{K}}{\sqrt{d_k}} Q_{\text{int},b}K_{\text{int},b}^T \\
U &= \text{softmax}(Y), \text{and} \\
A &= UV \approx s_{U} s_V U_\text{uint,b} V_\text{int,b}.
\end{align*}
$U$ is the attention-weight matrix, which is
the only matrix converted to \textit{unsigned} integers
because $\text{softmax}(QK^T)$ is positive.

\subsubsection{Number of Trained Threshold Scalars}
The Transformer architecture consists of a stack of $N=6$ encoder modules,
a stack of $N=6$ decoder modules, a linear-projection layer, and finally
a softmax layer.
Although Transformer Base has eight attention heads and
Transformer Big has 16 heads,
we use the same threshold scalar for each tensor operand
across all heads within the same module.
As a result, the number of threshold scalars is independent of the
number of attention heads.
Each encoder module consists of $D_\text{enc}=6$ \texttt{dense} layers
and $M_\text{enc}=2$ \verb+matmul+ layers.
Each decoder module consists of $D_\text{dec}=10$ \texttt{dense} layers
and $M_\text{dec}=4$ \verb+matmul+ layers.
Each \texttt{dense} layer needs just one threshold scalar for the
sole non-parameter input.
Each \texttt{matmul} layer requires two trained threshold scalars because neither
of the inputs is a parameter matrix.
We thus train
$N(D_\text{enc} + 2M_\text{enc} + D_\text{dec} + 2M_\text{dec})=168$ threshold scalars
among encoder and decoder modules.
The non-parameter tensor operand of the
final \texttt{dense} layer, whose weights are the embedding table,
uses one additional threshold scalar.
We therefore train 169 threshold scalars to make range-precision
trade-offs in the Transformer.

\section{Training Floating-Point Transformer Models}
To establish FP32 baselines, we trained both Transformer Base and Transformer Big
\cite{DBLP:conf/nips/VaswaniSPUJGKP17}
using the models provided in Tensor2Tensor v1.12 \cite{Vaswani2018}.
The key difference in the Transformer models in \cite{Vaswani2018}
and the original models in \cite{DBLP:conf/nips/VaswaniSPUJGKP17}
is that \cite{Vaswani2018} applied layer normalization
to the sentence representation matrix before computing attention weights.
This difference does not change the accuracy of Transformer Base and Transformer Big
but is beneficial for deeper Transformer models \cite{DBLP:conf/acl/WangLXZLWC19}.
We also used the training recipes in \cite{Vaswani2018},
which we outline in the next section.

\subsection{Training Data}
We used the default dataset in the Tensor2Tensor v1.12 
English-to-German translation task (\verb+translate_ende_wmt32k_packed+).
This dataset has 4.6 million sentence pairs
drawn from three WMT18 \cite{DBLP:conf/wmt/2018s} parallel corpora:
News Commentary V13, Europarl V7, and Common Crawl.
We used \verb+t2t-datagen+ to create a vocabulary table with 33288 subwords,
corresponding to about 142 million subwords in the training dataset.

\subsection{Hardware and and Hyperparameters}
We trained both the Transformer Base and Big models 
using eight-core Google Cloud TPUs.
Specifically, we trained Transformer Base on TPUv2-8 and Transformer Big on TPUv3-8,
both with a batch size of 2048 subwords per TPU core.
An epoch is therefore about 9000 steps.
We trained Transformer Base using the hyperparameter set \verb+transformer_tpu+
for 300,000 steps,
and we trained Transformer Big using \verb+transformer_big_tpu+ for 600,000 steps.
Unlike the hyperparameter sets for GPUs, 
those for TPUs use the Adafactor optimizer \cite{DBLP:conf/icml/ShazeerS18},
with $t_w = 10000$ steps for warm-up and an inverse square root decay schedule
such that the learning rate factor is $1/{\sqrt{\max{(t_w, t)}}}$.

\subsection{Post-FP32 Training}
Our goal is to convert floating-point tensors to integer tensors for
all \texttt{dense} and \texttt{matmul} layers.
We first tried simultaneously quantizing the dense-layer weight tensors
and learning the threshold scalars for non-parameter tensors.
However, this did not result in convergence.
Therefore, we tried a different approach, which, although simple,
achieved high BLEU scores without convergence problems.
In this approach, we fine-tuned parameter tensors and non-parameter tensors separately
over three to six epochs, the first three of which were mandatory and the
remaining three optional.

\begin{figure}[ht]
\vskip 0.2in
\begin{center}
\centerline{\includegraphics[width=\columnwidth]{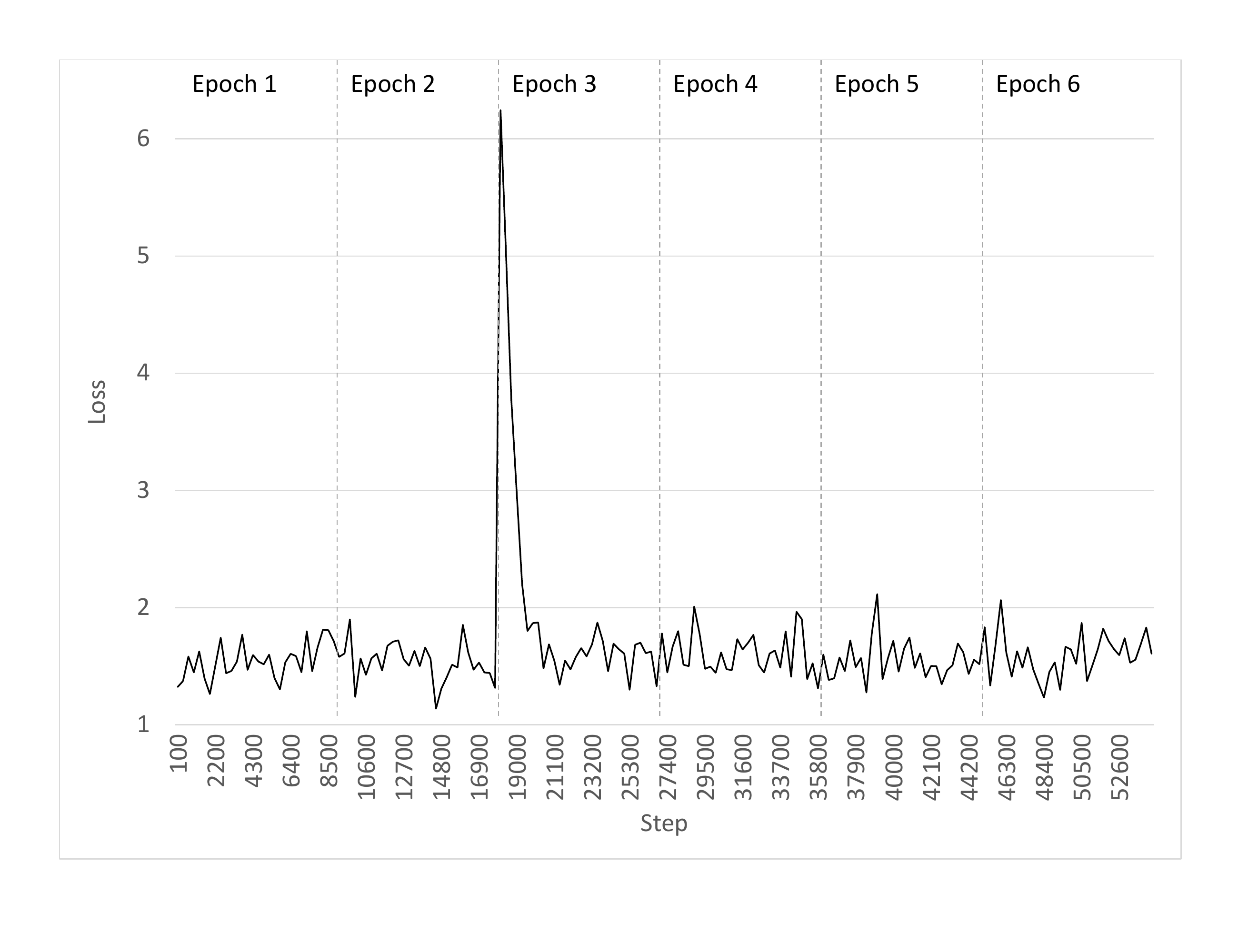}}
\caption{Loss after FP32 training of the Transformer Base model.
Dotted vertical lines delineate epochs.
The loss spike at the beginning of the third epoch is due to
unoptimized initial threshold scalars for non-parameter tensors.
In the third and fourth epochs, gradients with respect to parameters
stop propagating to allow threshold scalars to reduce loss,
in effect learning to make range-precision trade-offs.
}
\label{fig_transform_base_loss}
\end{center}
\vskip -0.2in
\end{figure}

\begin{enumerate}
\item In the first epoch, we converted weight and bias tensors to integers
using Eqs.\ \ref{eq:signed} to \ref{eq:sp_X}
while leaving non-parameter tensors in the floating-point format.
\item In the second epoch, we froze the integer parameter tensors and measured
the maximum absolute value of all FP32 non-parameter tensors that are
inputs to \texttt{dense} layers.
\item At the start of the third epoch, we initialized the threshold scalar of
each dense-layer non-parameter input,
while still using the same integer parameter tensor.
The training loss increased abruptly (Fig.\ \ref{fig_transform_base_loss}) but settled
quickly.
This pattern suggests that preserving the range for these tensors is suboptimal,
but clipping these tensors helps in minimizing the loss.
\item (Optional) The fourth epoch continued refining threshold scalars.
\item (Optional) In the fifth epoch, we froze threshold scalars and fine-tuned the integer parameters.
\item (Optional) The fine-tuning of integer parameters continued in this epoch.
\end{enumerate}

Fig.\ \ref{fig_transform_base_loss} illustrates the training loss
for six epochs after FP32-training.

\section{Experimental Results}
Using the validation dataset (\textit{newstest2013}),
we obtained integer weights from the checkpoint with the highest BLEU score
in the last two epochs of parameter fine-tuning.
We used \verb+t2t-bleu+ \cite{DBLP:journals/corr/abs-1804-00247}
to report the BLEU scores of FP32 models and integer models.
Scores from \verb+t2t-bleu+ are the same as those from
\verb+sacrebleu+ \cite{DBLP:journals/corr/abs-1804-08771}.
We used a beam size of 4 in beam search and length penalty 0.6, and
recorded both cased and uncased scores.
Table \ref{newstest2014_results}
shows BLEU scores for both FP32 models and INT8 models.
Unlike the BLEU scores in Table 2 in \cite{DBLP:conf/nips/VaswaniSPUJGKP17},
we did not apply checkpoint averaging to obtain Table \ref{newstest2014_results}.
Still, we obtained the same BLEU score for Transformer Base (27.3) and
a higher BLEU score for Transformer Big (29.2 as opposed to 28.4).

Because we used the same \texttt{tensor2tensor} code base,
validation dataset, and test dataset
as \cite{DBLP:journals/corr/abs-1906-00532} did,
our INT8 Transformer Base BLEU scores can be directly compared to theirs,
in which the FP32 baseline BLEU score was 27.68 and
the symmetric INT8 BLEU score was 27.30 (-1.4\%).
By constrast,
our INT8 Transformer Base BLEU score did not drop at all, either
cased or uncased.

Although \cite{DBLP:journals/corr/abs-1906-00532} did not report
results for Transformer Big,
our Transformer-Big INT8 BLEU scores can be compared to
\cite{DBLP:conf/aclnmt/Junczys-Dowmunt18},
in which the BLEU score dropped from 28.1 to 27.5 (-2.1\%).
By contrast, our Transformer Big INT8 BLEU score drop was only
0.7\% for cased and 0.3\% for uncased,
despite the higher baseline scores for the FP32 model at 29.2 cased
and 29.6 uncased.

To demonstrate the robustness of our approach,
we converted the same FP32 Transformer models to even lower
precision (INT6).
The rightmost column in Table \ref{newstest2014_results} shows
1.0 BLEU point drop for Transformer Base (cased and uncased),
1.3 BLEU point drop for Transformer Big uncased, and
1.4 BLEU point drop for Transformer Big cased.

To go beyond the \textit{newstest2014} dataset,
we reported FP32 and INT8 BLEU scores in Table \ref{newstest2015_2019} using
the same models from Table \ref{newstest2014_results}.
We observed that 10 of the 24 BLEU scores from the INT8 models in Tables \ref{newstest2014_results} and \ref{newstest2015_2019}
are either the same or higher than those from the FP32 models.

\begin{table*}
\begin{center}
\caption{FP32, INT8, and INT6 Transformer \textit{newstest2014} English-to-German translation BLEU scores.}
\label{newstest2014_results}
\begin{tabular}{l|*{4}{c}}
\toprule 
Model  & {FP32} &  {INT8} & {INT6}\\\midrule
Transformer Base Uncased & 27.8 & 27.8 & 26.8 \\ %300K FP32, 350K int8
Transformer Base Cased   & 27.3 & 27.3 & 26.3 \\ 
Transformer Big Uncased  & 29.6 & 29.5 & 28.3\\ % 599K FP32, 645K int8
Transformer Big Cased    & 29.2 & 29.0 & 27.8\\\bottomrule 
\end{tabular}
\end{center}
\end{table*}

\begin{table*}
\begin{center}
\caption{FP32 and INT8 Transformer \textit{newstest2015} to \textit{newstest2019}
English-to-German translation BLEU scores.}
\label{newstest2015_2019}
\begin{tabular}{l|*{4}{cc|}cc}
\toprule 
Model &
\multicolumn{2}{c|}{15} &
\multicolumn{2}{c|}{16} & \multicolumn{2}{c|}{17} &
\multicolumn{2}{c|}{18} & \multicolumn{2}{c}{19} \\
\null                    & FP32 & INT8 & FP32 & INT8 & FP32 & INT8 & FP32 & INT8 & FP32 & INT8\\\midrule
Transformer Base Uncased  & 30.2 & 30.4 & 35.1 & 34.9 & 28.4 & 28.4 & 42.3 & 42.1 & 38.5 & 38.1\\
Transformer Base Cased   & 29.7 & 29.9 & 34.5 & 34.3 & 27.8 & 27.8 & 41.8 & 41.7 & 38.1 & 37.8\\
Transformer Big Uncased  & 32.0 & 31.9 & 35.9 & 35.8 & 29.5 & 29.5 & 43.2 & 43.1 & 39.2 & 39.4\\
Transformer Big Cased    & 31.6 & 31.4 & 35.3 & 35.2 & 29.0 & 29.0 & 42.7 & 42.6 & 38.8 & 39.0
\\\bottomrule 
\end{tabular}
\end{center}
\end{table*}

\section{Conclusion}
We presented a stable integer quantization approach for Transformer
machine-translation models that
converts an existing FP32 model to an integer model by
alternately optimizing parameter tensors and non-parameter tensors in separate epochs.
Unlike the case in previous work, we applied this approach to
\textit{all} 133 matrix multiplications in
\textit{both} Transformer Base and Transformer Big.
For the English-to-German translation task on the \textit{newstest2014} test set,
our INT8 models achieved 99.3\% to 100\% BLEU scores relative to FP32 models.
To show the robustness of this approach, we extended it to INT6,
although perhaps only FPGAs can take advantage of this non-standard format.
In addition,
we presented BLEU scores from \textit{newstest2015} to \textit{newstest2019}
for the INT8 models to illustrate the usefulness of these models.
It is encouraging that 10 out of the 24 BLEU scores from our INT8 models are at least as high as
the scores from the FP32 models.
Since our quantization approach starts with an FP32 model,
we hypothesize that a principled method in selecting local minima
(roughly training checkpoints)
in the FP32 optimization landscape may yield even better results
\cite{NIPS2019_8524},
which is a topic for future research.

\bibliography{../my_bib/my_lib}

\bibliographystyle{named}

\end{document}